\documentclass[sn-mathphys]{sn-jnl}

\jyear{2025}%

\theoremstyle{thmstyleone}%
\usepackage{amsmath}  
\usepackage{amssymb}  % For additional math symbols
\usepackage{mathrsfs} 
\usepackage{bbm}
\usepackage{physics}
\usepackage{xcolor}

\usepackage{graphicx}
\usepackage{tabularx}
\usepackage[table]{xcolor}

\usepackage{comment}

\usepackage{subcaption}

\theoremstyle{thmstyletwo}%
\definecolor{lilla}{RGB}{220, 208, 237}
\definecolor{acquamarina}{RGB}{182, 228, 236}

\theoremstyle{thmstylethree}%

\begin{document}
%\linenumbers

\title[Challenging DINOv3 Foundation Model ...]{Challenging DINOv3 Foundation Model under Low Inter-Class Variability: A Case Study on Fetal Brain Ultrasound}

\author[1]{\fnm{Edoardo Conti}}
\author*[1]{\fnm{Riccardo} \sur{Rosati}}\email{r.rosati@staff.univpm.it}
\author*[1]{\fnm{Lorenzo} \sur{Federici}}\email{l.federici@pm.univpm.it}
\author[1]{\fnm{Adriano} \sur{Mancini}} 
\author[1]{\fnm{Maria Chiara} \sur{Fiorentino}} 
\affil[1]{\orgdiv{Department of Information Engineering}, \orgname{Università Politecnica delle Marche}, \country{Italy}}

%%==================================%%
%% sample for unstructured abstract %%
%%==================================%%

\abstract{
{\bf{Purpose:}} This study provides the first comprehensive evaluation of foundation models in fetal ultrasound (US) imaging under low inter-class variability conditions. While recent vision foundation models such as DINOv3 have shown remarkable transferability across medical domains, their ability to discriminate anatomically similar structures has not been systematically investigated. We address this gap by focusing on fetal brain standard planes—transthalamic (TT), transventricular (TV), and transcerebellar (TC)—which exhibit highly overlapping anatomical features and pose a critical challenge for reliable biometric assessment.

{\bf{Methods:}} To ensure a fair and reproducible evaluation, all publicly available fetal ultrasound datasets were curated and aggregated into a unified multicenter benchmark, FetalUS-188K, comprising more than 188,000 annotated images from heterogeneous acquisition settings. DINOv3 was pretrained in a self-supervised manner to learn ultrasound-aware representations. The learned features were then evaluated through standardized adaptation protocols, including linear probing with frozen backbone and full fine-tuning, under two initialization schemes: (i) pretraining on FetalUS-188K and (ii) initialization from natural-image DINOv3 weights.

{\bf{Results:}} 
Models pretrained on fetal ultrasound data consistently outperformed those initialized on natural images, with weighted F1-score improvements of up to 20\%. Domain-adaptive pretraining enabled the network to preserve subtle echogenic and structural cues crucial for distinguishing intermediate planes such as TV.

{\bf{Conclusion:}} 
Results demonstrate that generic foundation models fail to generalize under low inter-class variability, whereas domain-specific pretraining is essential to achieve robust and clinically reliable representations in fetal brain ultrasound imaging.

}


\maketitle

% how to make ...
\section{Introduction}
\label{sec:intro}

Fetal standard planes—such as abdominal, brain, and cardiac views—are fundamental in prenatal ultrasound (US) imaging, as they provide the basis for fetal biometry and enable the detection of conditions like growth restriction and congenital anomalies \cite{migliorelli2024use, fiorentino2022review}. Traditionally, identifying these planes has relied on the expertise of trained sonographers. However, the manual interpretation of US images is inherently subjective, which may compromise the consistent and accurate localization of standard planes. This has motivated the adoption of deep learning (DL) techniques to automate plane identification, thereby enhancing both efficiency and precision in fetal biometry \cite{bai2024psfhs}.
Early DL approaches for standard plane detection predominantly relied on single-center datasets, which, while useful for proof-of-concept studies, suffer from limited generalizability due to restricted population diversity and site-specific imaging protocols \cite{kiserud2017world}. The release of the first publicly available datasets, such as those introduced by \cite{burgos2020evaluation} and subsequently employed in \cite{krishna2024standard,migliorelli2024use,krishna2023automated}, marked an important step toward reproducible research in this domain. However, these datasets still originated from limited clinical settings, raising concerns about model robustness when deployed across diverse healthcare environments. Recognizing these limitations, the research community has progressively shifted toward multicenter studies \cite{sendra2023generalisability, krishna2024deep, fiorentino2025contrastive}, which aggregate data from multiple hospitals and populations to better capture the heterogeneity of real-world clinical practice. More recently, this evolution has culminated in the development of generalizable foundation models \cite{ambsdorf2025general}, which leverage large-scale pretraining on diverse vision tasks to achieve transferable representations for medical imaging. While these models demonstrate robust performance across standard planes with high inter-class variability—such as abdominal circumference and four-chamber heart views, where anatomical structures are distinctly different—a critical gap remains in their evaluation on more challenging scenarios. Specifically, their ability to discriminate between anatomically similar planes has not been systematically investigated. Brain planes such as transventricular (TV), transcerebellar (TC), and transthalamic (TT) views exemplify this challenge: they share overlapping structural features, including similar echogenic patterns and the presence of key landmarks like the midline, lateral ventricles, and posterior fossa structures. This low inter-class variability makes their differentiation particularly difficult, yet accurate classification is essential for reliable biometric measurements—such as head circumference and biparietal diameter—and for the detection of neurological anomalies \cite{gallery2020isuog}.

In this work, we address this gap by conducting the first comprehensive evaluation of foundation models on fetal brain planes with low inter-class variability. To ensure a fair and reproducible assessment that reflects real-world clinical diversity, we aggregate all publicly available fetal US datasets from the literature, creating the largest and most heterogeneous benchmark to date for this specific task. Our contributions are fourfold: 
(i) we present, to the best of our knowledge, the first application of DINOv3 to US imaging, considering its potential to better capture domain-specific characteristics compared to previous versions; (ii) we establish a unified and curated multi-center benchmark (FetalUS-188K) that consolidates and harmonizes fragmented public datasets through a standardized cleaning and filtering procedure, enabling rigorous comparison across methods and promoting reproducibility in future research; (iii) we provide the first systematic analysis of foundation model performance on anatomically similar brain planes, revealing critical limitations in handling subtle inter-class differences that are not apparent when evaluating on standard, high-variability plane classifications; and (iv) we demonstrate that while foundation models exhibit strong feature extraction capabilities, targeted fine-tuning strategies are necessary to achieve clinically acceptable discrimination between TV, TC, and TT views.

%%%%%
\begin{figure}[tbp]
    \centering
    \includegraphics[width=0.99\textwidth]{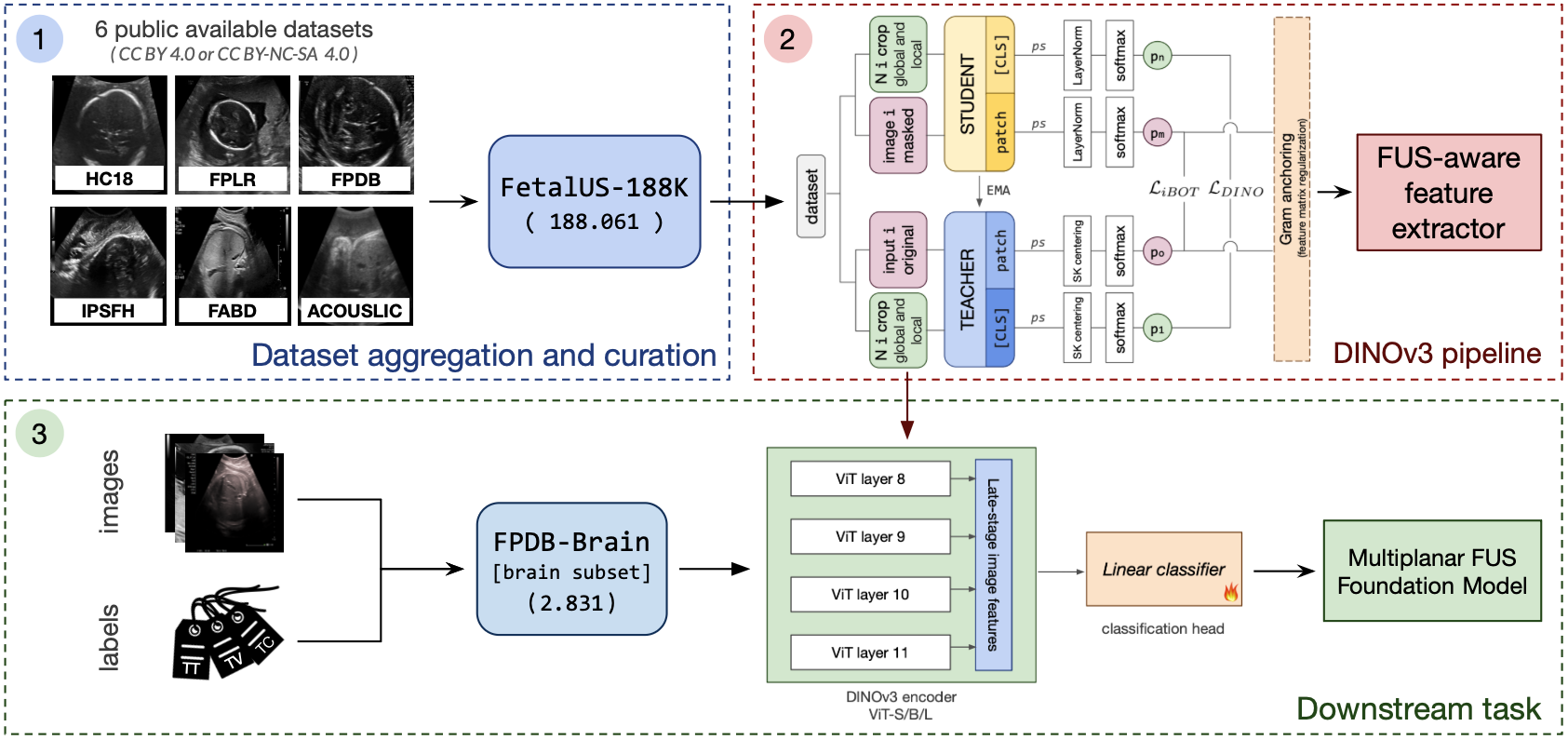}
    \caption{Overview of the proposed pipeline. (1) All publicly available fetal ultrasound datasets are aggregated and curated to build the FetalUS-188K dataset, ensuring heterogeneous and diverse ultrasound acquisitions. (2) A self-supervised training strategy based on the DINOv3 pipeline is employed to learn a feature extractor specifically aware of fetal ultrasound characteristics. (3) The learned representations are transferred to the downstream task of fetal brain standard plane classification. }
    \label{fig:arch}
\end{figure}
%%%%%%%%%

\section{Method}
\begin{comment}
--> questo è DINOv2
To assess the capability of foundation models to distinguish fetal brain planes that exhibit low inter-class variability, we adopt DINOv3 as our baseline \cite{liu2025does}. DINOv3 is a state-of-the-art self-supervised framework based on Vision Transformers (ViTs) and has shown strong performance in medical imaging scenarios. Its distilled representation learning supports robust feature extraction in the presence of domain shifts and noise, which are common in US acquisitions.

The overall framework is illustrated in Figure \ref{fig:arch}. We rely on the self-supervised objective of DINOv3, which enforces representation consistency across different augmented views of the same image. Given an US image $\mathbf{x}$, two augmented views $\mathbf{x}1$ and $\mathbf{x}2$ are generated through stochastic transformations. The student network $f\theta$ and the teacher network $f{\theta'}$ extract features from the two views, and the student is optimized to align its predictions with those of the teacher:

\begin{equation}
\mathcal{L}{\text{pre}} = -\sum{i=1}^{2} \sum_{j=1}^{2} \mathbb{1}{i \neq j} \cdot P{\theta'}(\mathbf{x}i)^\top \log P\theta(\mathbf{x}_j)
\end{equation}

\noindent where $P_\theta$ and $P_{\theta'}$ are softmax-normalized network outputs, controlled by a temperature parameter $\tau$. The teacher parameters are updated through an exponential moving average: $\theta' \leftarrow m\theta' + (1-m)\theta$, with $m \in [0,1]$ acting as a momentum coefficient.
\end{comment}

To assess the capability of foundation models in scenarios characterized by low inter-class variability, such as fetal brain planes, we have chosen DINOv3 as our baseline model \cite{liu2025does}. DINOv3, a state-of-the-art self-supervised framework based on Vision Transformers (ViTs), introduces significant improvements over its predecessor DINOv2. Its backbone enables the extraction of scalable and cross-domain features, incorporating techniques like multi-crop strategies, Gram anchoring, and patch-level latent reconstruction. This is crucial where challenges like domain shifts and noise are common, as seen in US acquisitions.

The overall architecture is showed in Figure \ref{fig:arch}. DINOv3 is based on a composite self-supervised objective, combining a global consistency loss with a patch-level latent reconstruction loss. Given any US image $ \mathbf{x} $, multiple views (two global crops $ \mathbf{x}1 $, $ \mathbf{x}2 $ and several local crops) are obtained via stochastic transformations. Both the student network $f\theta$ and the teacher network $f{\theta’}$ extract features from each view. The student is then optimized to align its global outputs with those of the teacher, while simultaneously reconstructing patch-level latent representations from masked local crops. A preliminary training objective is:

\[
\mathcal{L}_{\mathrm{Pre}} = \mathcal{L}_{\text{DINO}} + \mathcal{L}_{\text{iBOT}} + 0.1 \cdot \mathcal{L}_{\mathrm{Koleo}}
\]

where $\mathcal{L}_{\text{DINO}}$ enforces alignment between global student and teacher predictions, $\mathcal{L}_{\text{iBOT}}$ is a patch-level latent reconstruction loss across masked local views, and $\mathcal{L}_{\mathrm{Koleo}}$ encourages uniform spread of features in the embedding space. Outputs of all networks are softmax-normalized, controlled by temperature parameter $\tau$, while multi-crop heads use LayerNorm for stabilization.

After the initial training, DINOv3 introduces a Gram anchoring phase, where the model is further trained to preserve local spatial correspondences by regularizing the Gram matrix similarity between patch embeddings at different checkpoints. This process effectively maintains dense feature quality at scale. The teacher parameters are updated using an exponential moving average: $\theta’ \leftarrow m\theta’ + (1-m)\theta$, where $m \in [0,1]$ represents the momentum.

Unsupervised pre-training is performed on the entire multi-center dataset, including all fetal US views. This strategy enables the model to learn generalizable US-specific representations, capturing key characteristics such as speckle patterns, acoustic shadowing, and variations in anatomical textures. The resulting pretrained encoder is then specialized for fetal brain plane classification in the downstream task.

\subsection{Task Adaptation: Fetal Brain Planes}
\label{sec:method}
% -> to assess us pretrianing sono state valutati i tre tagli backbone dinov3: vit-b/16, vut-l/\6 e vit-s -> di questi si valuta LP. Allo stesso modo è valutato la capacità di genreralizzazione dei pesi dinov3 considerando sempre il Vit-b/16.
% -> un'altra valyutazione è stata fatta sul numero dei parametri, 

To evaluate the quality of the visual representations learned through fetal US-based pretraining of DINOv3, we employ a standardized, lightweight adaptation protocol that minimizes task-specific parameters as much as possible. This ensures that performance reflects the representational strength of the frozen backbone rather than the adaptation head. Specifically, linear probing (LP) is performed by freezing the pretrained backbone and training a shallow linear classifier on top of the extracted features. Following the evaluation protocol in ~\cite{oquab2023dinov2, simeoni2025dinov3}, multiple classifiers are trained under different configurations, and the best model is selected according to validation accuracy.

LP evaluations are conducted under two initialization schemes: (i) pretraining on fetal US data (ours) and (ii) initialization from publicly released DINOv3 weights~\cite{simeoni2025dinov3}. The transferability of the learned representations is further assessed by fully fine-tuning (FT) the entire network under the same two initialization settings.

%%%%%%%%%%%%%%%%%%%%
\begin{table}[t]
    \centering
    %\tiny
    \caption{Datasets information including country of acquisition, ultrasound device type, specific task, and number of annotated samples per class. BPE = biometry parameter estimation; SPD = standard plane detection; ASA = anatomical structure analysis.}
    \label{tab:ultrasound_data}
    \renewcommand{\arraystretch}{1.2}
    \begin{tabularx}{\textwidth}{ccXcX}
        \hline
        \textbf{Name} & \textbf{Country} & \textbf{Device} & \textbf{Task} & \textbf{Number of Samples} \\
        \hline
        HC18 & Netherlands & Voluson E8 and Voluson 730 &
        BPE &
        Brain (999) \\
        FPDB & Spain & Voluson E6, Voluson S8, Voluson S10, Aloka &
        SPD &
        Abdomen (353), Femur (516), Thorax (1.058), Maternal cervix (981) \\
        FPLR & Africa & Mindray DC-N2, Voluson P8, ACUSON X600, EDAN DUS 60, Voluson S8 &
        SPD &
        Abdomen (100), Brain (100), Femur (100), Thorax (75) \\
        FABD & Brazil & Voluson 730, Philips-EPIQ Elite &
        ASA &
        Abdomen (1.100) \\
        IPSFH & China & ObEye system &
        BPE & PS and FH (4.000) \\
        ACOUSLIC & Africa & MicrUs Pro-C60S &
        BPE & Abdomen (178.679) \\
        \hline
    \end{tabularx}
\end{table}

%%%%%%%%%%%%%%%%%

%%%%%%%%%%%%%%%%%%%%%%%%%

\section{Experiments}
\label{setting}

\subsection{ FetalUS-188K Dataset}
\label{dataset}

To ensure a robust and generalizable evaluation, we aggregate all currently available public fetal US datasets into a multi-center benchmark (FetalUS-188K) that encompasses diverse acquisition settings, probe characteristics, population demographics, and different downstream tasks, ranging from biometric parameter estimation to standard plane detection and anatomical structure analysis (see Table \ref{tab:ultrasound_data}).

Our aggregated benchmark comprises six complementary datasets:
i) HC18 Challenge dataset\footnote{\url{https://zenodo.org/records/1327317}} (HC18), originally developed for head circumference estimation, providing TV and TT brain views;
ii) Fetal Planes DB\footnote{\url{https://zenodo.org/records/3904280}} (FPDB), a large multi-class dataset covering key diagnostic views including abdomen, brain, femur, thorax, and maternal cervix;
iii) Low-Resolution Fetal Planes Africa\footnote{\url{https://zenodo.org/records/7540448}} (FPLR), collected in resource-limited clinical environments, introducing substantial domain variability in imaging quality and patient demographics;
iv) Pubic Symphysis and Fetal Head\footnote{\url{https://zenodo.org/records/7851339}} (IPSFH), symphysis pubis and fetal head from ITU images collected by clinical radiologists for head segmentation and angle of progression;
v) Fetal Abdominal Structures dataset\footnote{\url{https://data.mendeley.com/datasets/4gcpm9dsc3/1}} (FABD), providing abdominal planes with pixel-wise annotations of primary abdominal structures;
vi) Abdominal Circumference Operator-agnostic UltraSound measurement in Low-Income Countries dataset\footnote{\url{https://zenodo.org/records/12697994}} (ACOUSLIC), comprising blind-sweep 2D prenatal abdominal US sequences for anatomical landmark localization and biometric estimation. A substantial portion of the frames in ACOUSLIC represent transitional segments or contain negligible visual information. To ensure data consistency and prevent training on non-informative samples, an automatic filtering procedure was applied using an empirically defined threshold to exclude dark or low-informative ultrasound frames. This reduced the dataset from 252{,}840 to 178{,}679 valid frames.

The resulting aggregated dataset serves as the training foundation for our study, enabling the model to learn rich and transferable fetal anatomical representations across diverse clinical domains. For the downstream evaluation task, consisting of TC, TT, and TV plane classification, we adopt the standardized subset introduced by Burgos-Artizzu et al. \cite{burgos2020evaluation}, following the same train/validation/test split provided by the authors.

\subsection{Implementation details}
The experimental configuration is identical across all evaluated models. The initialization is performed from DINOv3 pretrained weights, either obtained from our FetalUS-188K dataset or from the original LVD-1689M dataset \cite{simeoni2025dinov3}. We adopt ViTs as backbone architectures, evaluating three model variants—ViT-S/16, ViT-B/16, and ViT-L/16—which differ in network capacity and depth. The suffix “/16” denotes a patch size of 16×16 pixels. %meaning that each input image is divided into non-overlapping patches of this size before being linearly embedded and processed by the transformer layers. 
In all configurations, visual features are extracted from the last four transformer blocks for downstream adaptation.

All models are trained for 150 epochs using a batch size of 16 and 224×224 pixel input resolution. Optimization uses a cosine annealing learning rate scheduler with weight decay set to $1e^{-2}$ and an early stopping criterion. The best-performing models are selected among those tested across the hyperparameter search space, choosing the configuration achieving the highest validation accuracy. The search space includes several learning rates ranging from 1e$^{-5}$ to 1e$^{-1}$, different numbers of transformer blocks (indicating how many of the last ViT layers are used to extract and concatenate features for the linear classifier), and multiple loss functions. The tested losses comprise: the standard cross-entropy; the focal loss, which reduces the impact of easy examples while emphasizing harder ones,; two focal variants with different focusing parameters ($\gamma = 1$ and $\gamma = 3$), and the label smoothing cross-entropy, which helps prevent overfitting by introducing noise into the target labels.

%%%%%%%%%%%%%%%%%
\begin{comment}
\begin{table}[t]
    \centering
    \caption{TODO}
    \label{tab:exp_settings}
    \begin{tabularx}{\textwidth}{lc}
        \hline
        \multicolumn{2}{c}{\textbf{Settings}} \\
        \hline
        Pre-training & FetalUS-188K \textit{ or } LVD-1689M \\
        DINOv3 backbone & ViT \textit{ or } ConvNeXT \\
        DINOv3 feature layers & $[8, 9, 10, 11]$ \\
        Epochs & $150$ \\
        Batch size & $16$ \\
        Wights decay & $1e^{-2}$ \\
        Image resolution & $224x224$ \\
        Scheduler & cosine annealing \\
        Classes & [TT, TC, TV] \\
        \hline
    \end{tabularx}
\end{table}
\end{comment}
%%%%%%%%%%%%%%%

%\begin{table}[t]
%\centering
%\caption{Hyperparameter search space}
%\begin{tabular}{l c}
%\toprule
%\textbf{Hyperparameter} & %\textbf{Values} \\
%\midrule
%Learning rates & [1e$^{-5}$, 1e$^{-4}$, 1e$^{-3}$, 1e$^{-2}$, 1e$^{-1}$] \\
%N\_blocks & [1, 2, 4] \\
%Loss functions & [cross-entropy, focal, focal\_g1, focal\_g3, label smoothing] \\
%\bottomrule
%\end{tabular}
%\label{tab:hparams_search}
%\end{table}

%%%%%%%%%%

To reduce overfitting and enhance robustness to real acquisition variability, we apply a comprehensive augmentation pipeline. Geometric transformations include horizontal flipping ($p = 0.5$), affine transformations with $\pm 8\%$ translation and $\pm 12^\circ$ rotation ($p = 0.4$), and elastic deformations ($p = 0.15$), reflecting probe motion and anatomical changes during scanning. Intensity augmentations mimic differences in US equipment and operator settings through brightness and contrast adjustments ($\pm 15\%, p = 0.35$), gamma correction ($p = 0.25$), CLAHE enhancement ($p = 0.25$), and Gaussian noise injection ($p = 0.15$). %All transformations are combined with ImageNet normalization for compatibility with DINOv3 weights, while the validation split employs only normalization.

All experiments are implemented in PyTorch, building on the official DINOv3 repository. Data augmentation is performed using the Albumentations library to ensure reproducibility. The full codebase is publicly available\footnote{https://github.com/edoardo-conti/fetalus-fm}
. Experiments were run on the CINECA HPC infrastructure, leveraging nodes equipped with 32-core Intel Xeon CPUs, 512 GB RAM, and NVIDIA A100 GPUs with 64 GB of memory.

\subsection{Performance metrics}
Model performance is assessed using both quantitative and qualitative evaluation criteria. For the quantitative analysis, we compute the confusion matrix and the weighted F1-score, which represents the average of per-class F1 scores weighted by the number of instances in each class. In addition, per-class F1 scores are also reported where the averaging method corresponds to the macro formulation. For the qualitative analysis, we perform a Principal Component Analysis (PCA) on the extracted feature representations and visualize the first three principal components, along with their RGB composite.

%%%%%%%%%%%%%%%%%%%%%%%%%%%%%%%%%%%%%%%%%%
% Results CM LP

\begin{figure}[t]
    \centering

    % Row 1
    \begin{subfigure}{0.32\textwidth}
        \centering
        \includegraphics[width=\textwidth]{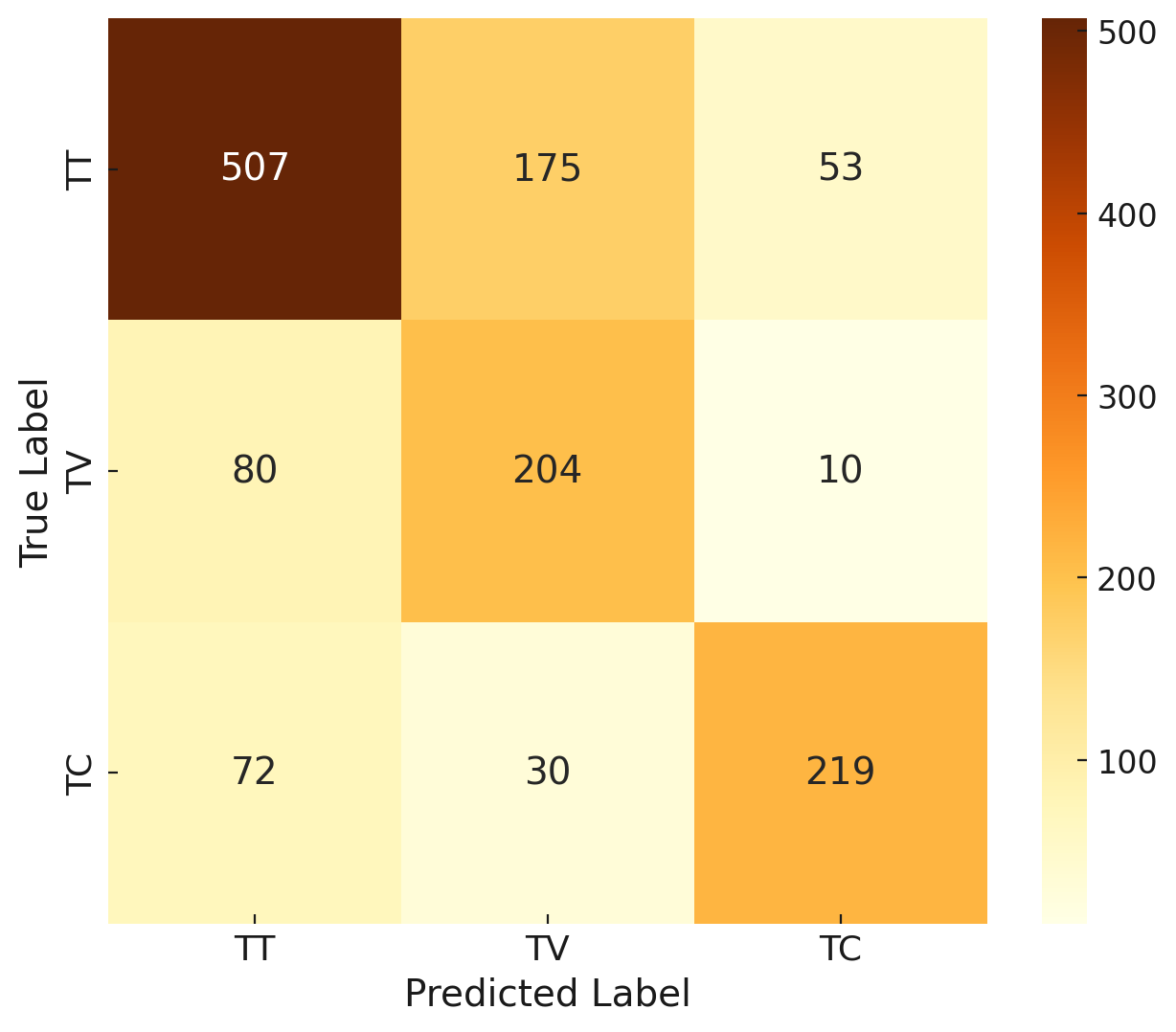}
        \caption{ViT-S FetalUS-188K}
        \label{fig:cm_vits_fus}
    \end{subfigure}
    \hfill
    \begin{subfigure}{0.32\textwidth}
        \centering
        \includegraphics[width=\textwidth]{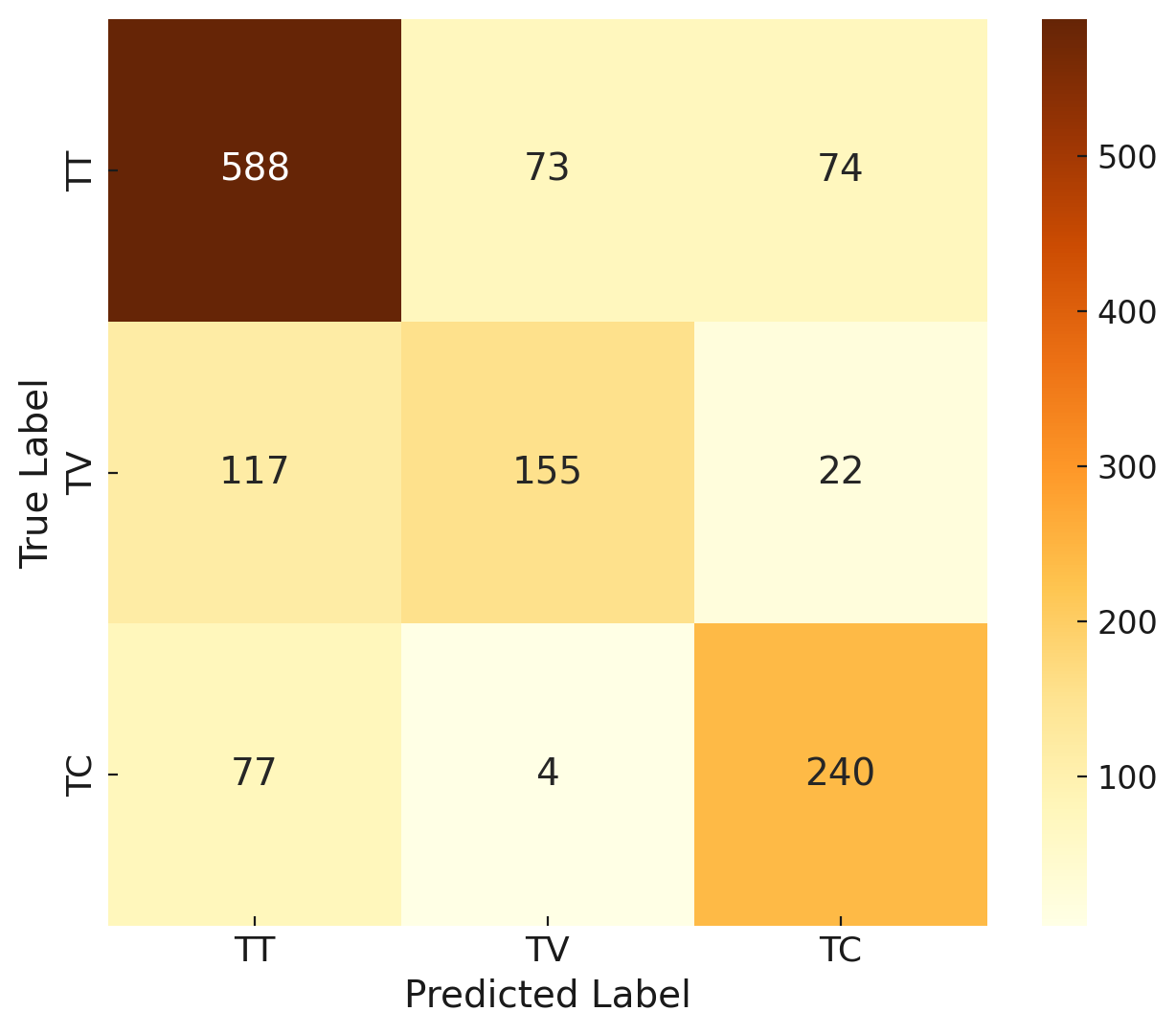}
        \caption{ViT-B FetalUS-188K}
        \label{fig:cm_vitb_fus}
    \end{subfigure}
    \hfill
    \begin{subfigure}{0.32\textwidth}
        \centering
        \includegraphics[width=\textwidth]{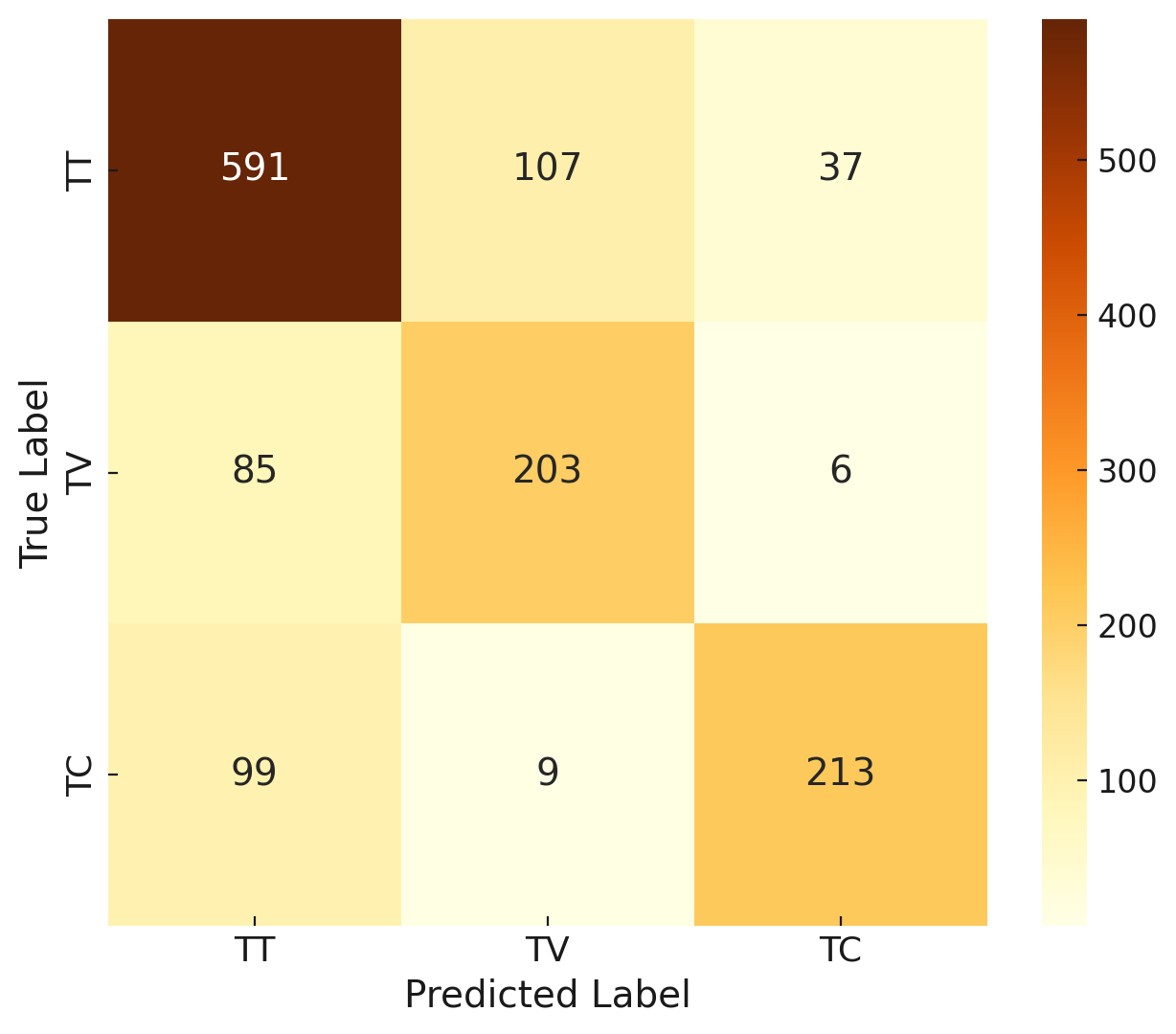}
        \caption{ViT-L FetalUS-188K}
        \label{fig:cm_vitl_fus}
    \end{subfigure}
    
    \vspace{0.35cm}

    % Row 2
    \begin{subfigure}{0.32\textwidth}
        \centering
        \includegraphics[width=\textwidth]{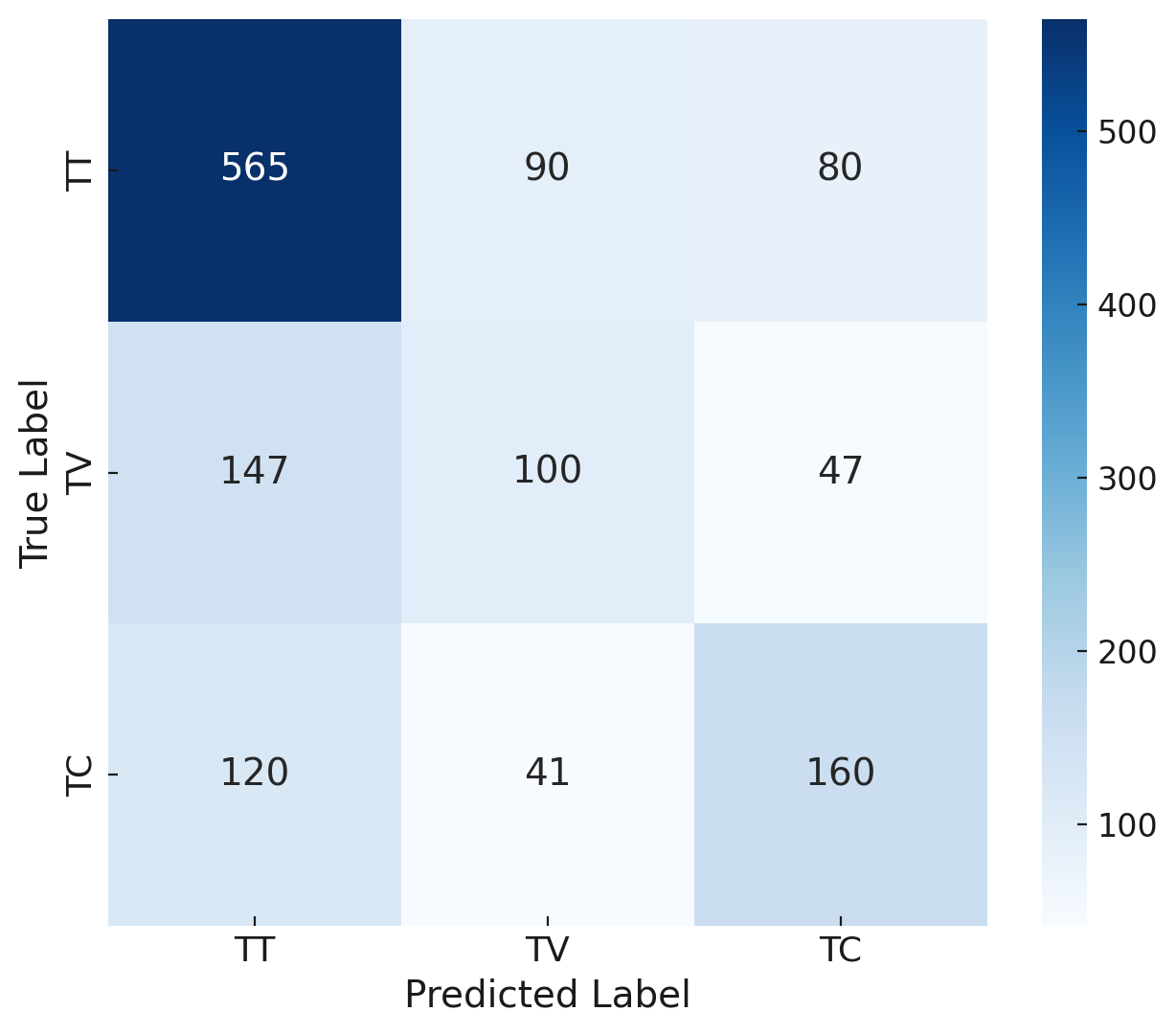}
        \caption{ViT-S LVD-1689M}
        \label{fig:cm_vitb_lvd}
    \end{subfigure}
    \hfill
    \begin{subfigure}{0.32\textwidth}
        \centering
        \includegraphics[width=\textwidth]{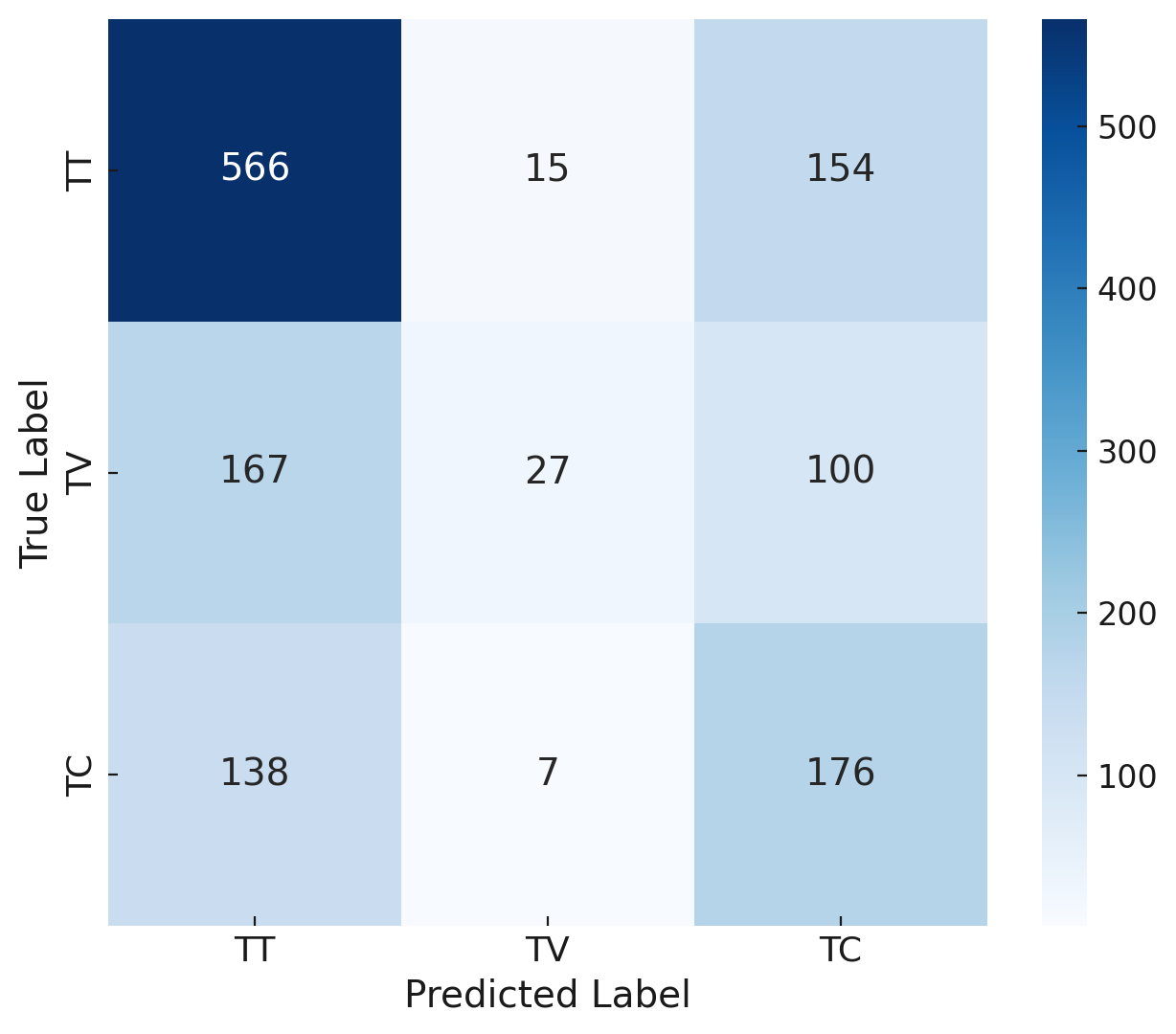}
        \caption{ViT-B LVD-1689M}
        \label{fig:cm_vitb_ftlvd}
    \end{subfigure}
    \hfill
    \begin{subfigure}{0.32\textwidth}
        \centering
        \includegraphics[width=\textwidth]{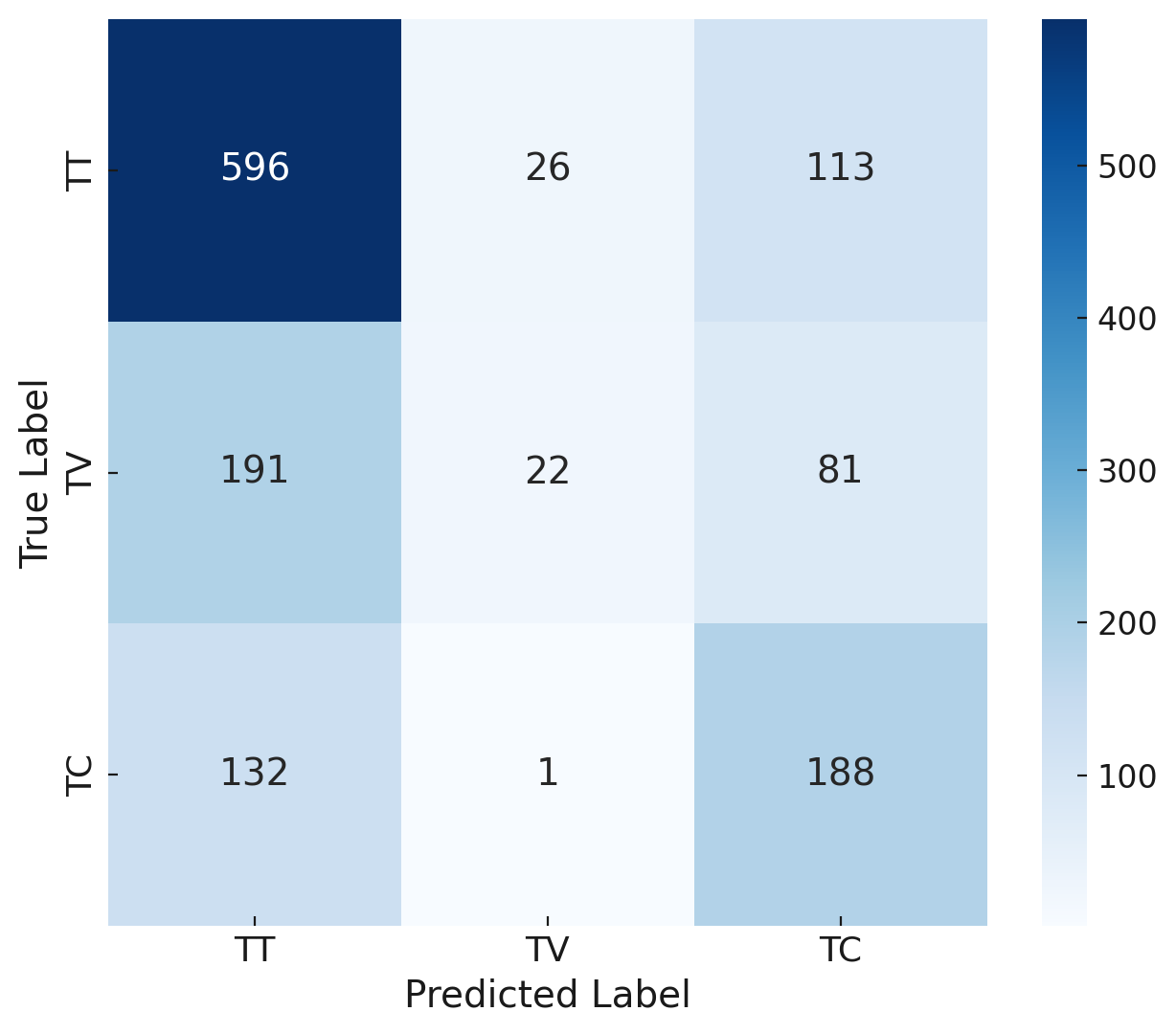}
        \caption{ViT-L LVD-1689M}
        \label{fig:cm_vits_ftlvd}
    \end{subfigure}

    \caption{Confusion matrices for TT, TV, TC obtained using linear probing on different ViT architectures. Top row: FetalUS-188K pretrained weights; bottom row: LVD-1689M ones.}
    \label{fig:cm_lp}
\end{figure}
%%%%%%%%%%%%%%%%%%%%%%%%%%%%%%%%%%%%%%%%%%

%%%%%%%%%%%%%%%%%%%%%%%%%%%%%%%%%%%%%%%%%%
% Results CM Fine tuning

\begin{figure}[t]
    \centering

    % Row 1
    \hspace{0.04\textwidth}
    \begin{subfigure}{0.32\textwidth}
        \centering      \includegraphics[width=\textwidth]{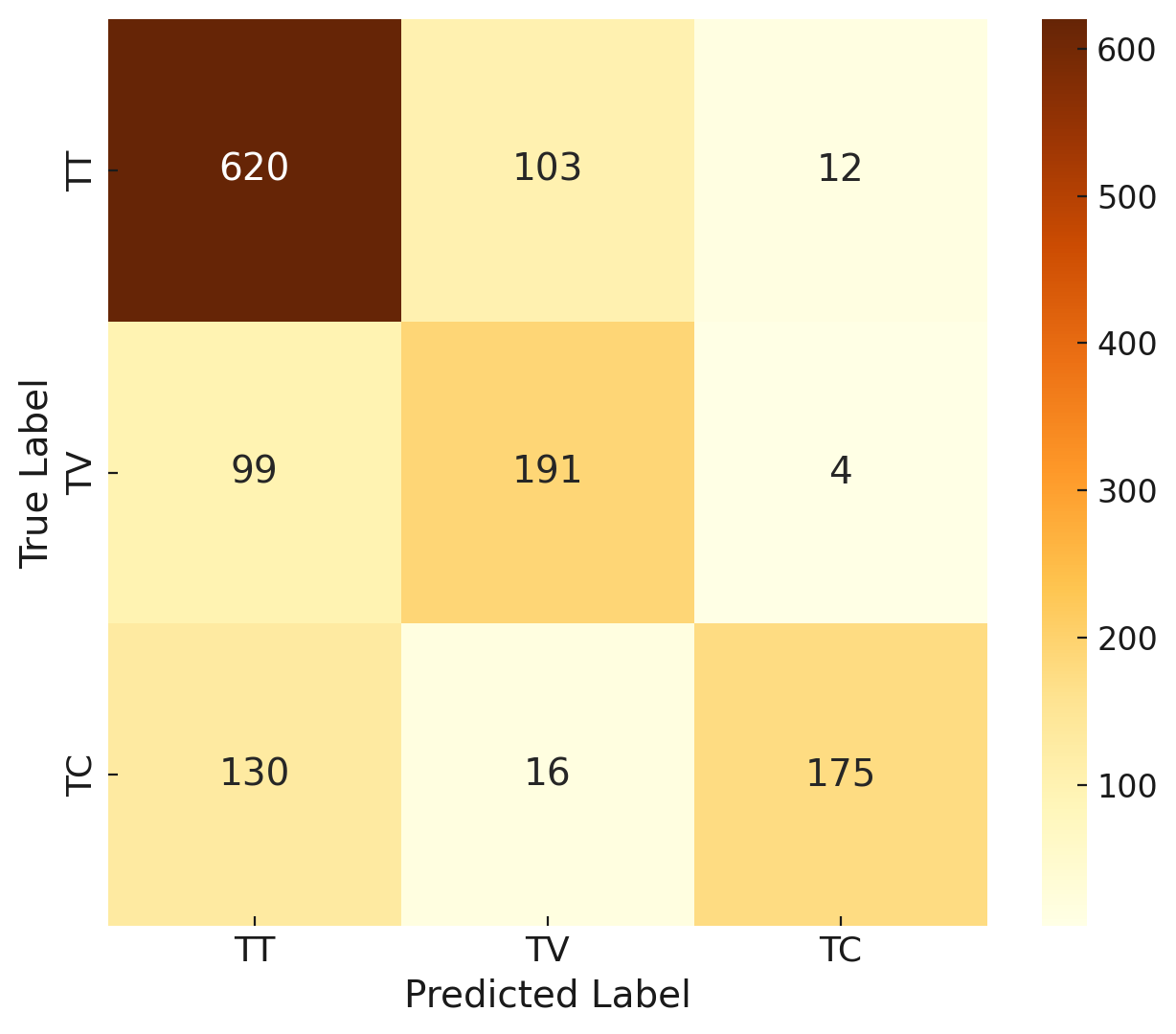}
        \caption{ViT-B FetalUS-188K}
        \label{fig:cm_vitl_fus}
    \end{subfigure}
    \begin{subfigure}{0.32\textwidth}
        \centering
        \includegraphics[width=\textwidth]{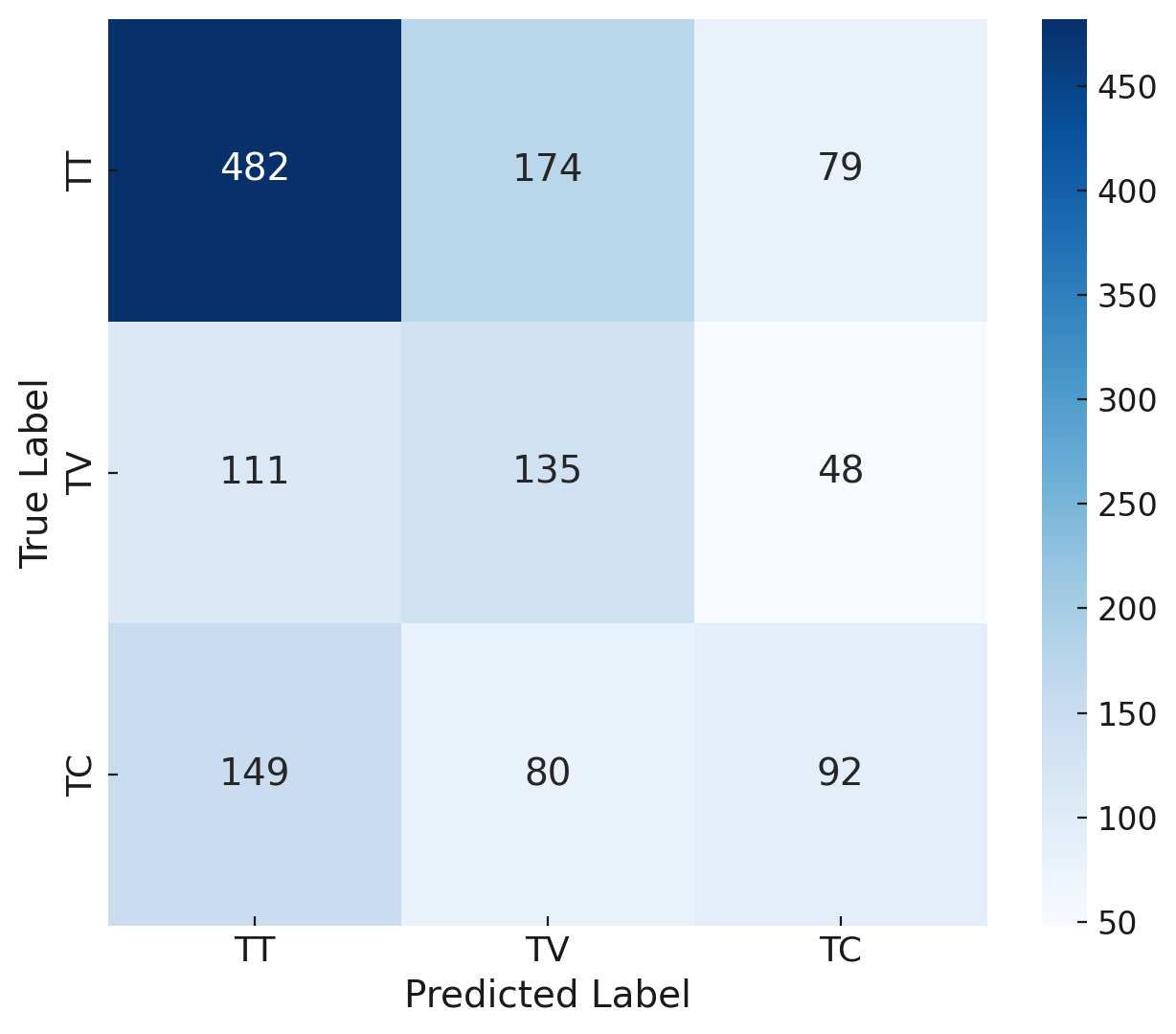}
        \caption{ViT-B LVD-1689M}
        \label{fig:cm_vitb_fus}
    \end{subfigure}

    \caption{Confusion matrices for TT, TV, TC obtained using full fine-tuning on ViT-B/16 architecture. Left: FetalUS-188K pretrained weights; right: LVD-1689M pretrained ones.}
    \label{fig:cm_all}
\end{figure}
%%%%%%%%%%%%%%%%%%%%%%%%%%%%%%%%%%%%%%%%%%

%%%%%%%%%%%%%%%%%%%%
\begin{table}[t]
\caption{Linear probing performance of different ViT architectures on FetalUS-188K and LVD-1689M pretraining, showing weighted and per-class F1-scores for TT, TV, and TC planes, achieved with best configuration reported as [N\_blocks, LR, Loss].}
\label{tab:macro-f1}
\begin{tabularx}{\linewidth}{lccc} % ccp{3.5cm}c
\toprule
\textbf{Dataset} & \textbf{Model} & \textbf{Best configuration} & \textbf{Weighted F1-score} \\
\midrule
\multirow{3}{7em}{FetalUS-188K} & ViT-S/16 & [4, $0.00625$, label\_smooth] & 0.69 [0.71, 0.60, 0.72] \\
 & ViT-B/16 & [4, $0.00063$, focal\_g1] & 0.72 [0.78, 0.59, 0.73] \\
 & \textbf{ViT-L/16} & \textbf{[4, 0.00625, focal\_g3]} & \textbf{0.74 [0.78, 0.66, 0.74]} \\
\midrule
\multirow{3}{7em}{LVD-1689M} & ViT-S/16 & [4, $0.00625$, focal] & 0.60 [0.72, 0.38, 0.52] \\
 & ViT-B/16 & [4, $0.00625$, focal\_g3] & 0.53 [0.70, 0.16, 0.47] \\
 & ViT-L/16 & [4, $0.00625$, focal\_g3] & 0.55 [0.72, 0.13, 0.53] \\
\bottomrule
\end{tabularx}
\end{table}
%%%%%%%%%%%%%%%%%%%%

%%%%%%%%%%%%%%%%%%%%
\begin{table}[t]
\caption{Fine-tuning performance of ViT-B/16 on FetalUS-188K and LVD-1689M pretraining, showing weighted and per-class F1-scores for TT, TV, and TC planes, achieved with best configuration reported as [N\_blocks, LR, Loss].}
\label{tab:macro-f1-ft}
\newcolumntype{Y}{>{\centering\arraybackslash}X}
\begin{tabularx}{\linewidth}{lYc}
\toprule
\textbf{Dataset} & \textbf{Best configuration} & \textbf{Weighted F1-score} \\
\midrule
LVD-1689M & [2, $0.00063$, focal] & 0.52 [0.65, 0.39, 0.34] \\
\textbf{FetalUS-188K} & \textbf{[4, 0.00625, focal\_g1]} & \textbf{0.73 [0.78, 0.63, 0.68]} \\
\bottomrule
\end{tabularx}
\end{table}
%%%%%%%%%%%%%%%%%%%%

%%%%%
\begin{figure}[tbp]
    \centering
    \includegraphics[width=0.70\textwidth]{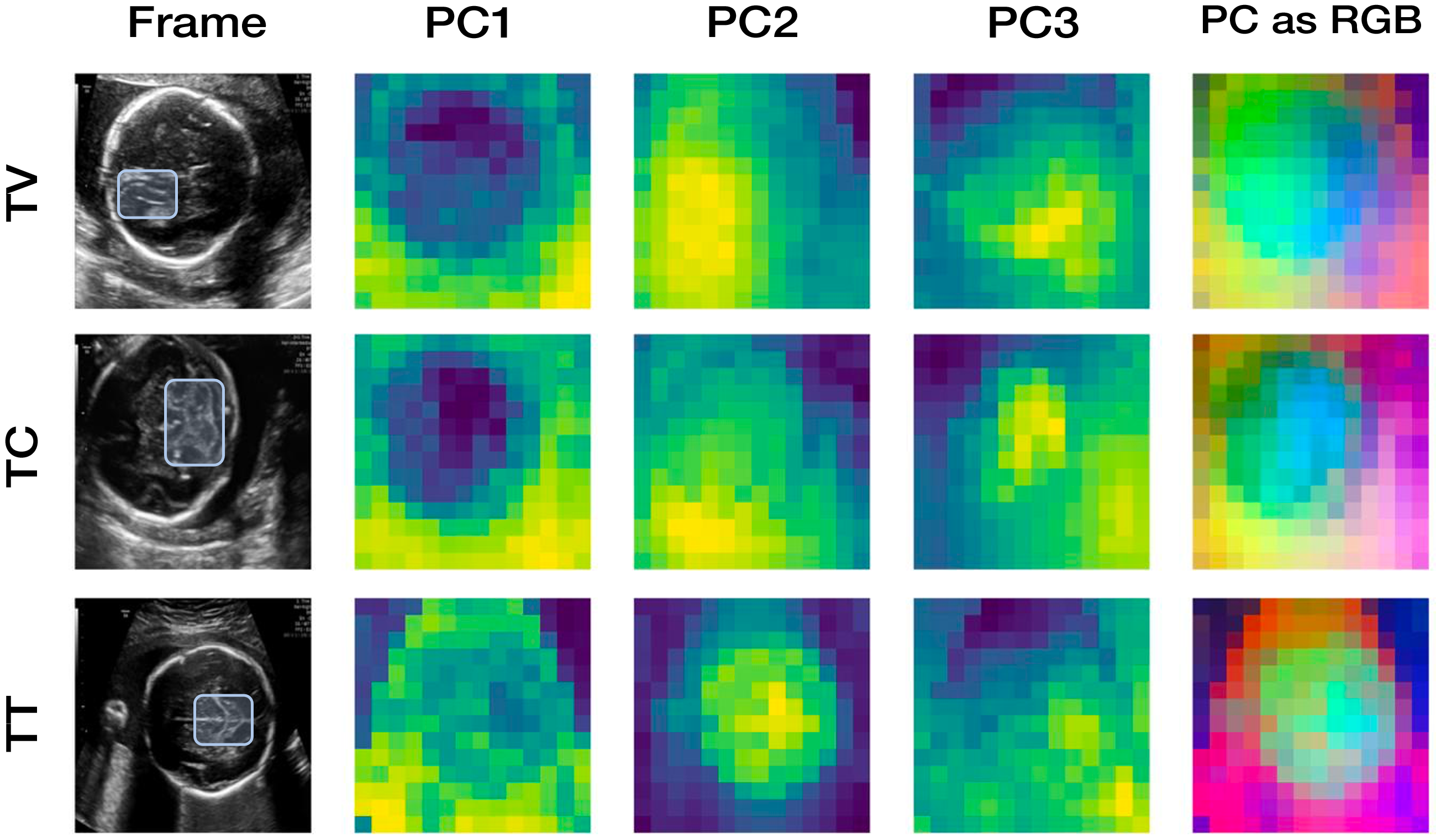}
    \caption{Principal component visualization from FetalUS-188K pretrained DINOv3 features. Each row corresponds to a fetal brain standard plane (TV, TC, TT), and columns show the first three principal components (PC1–PC3) and their RGB composite. }
    \label{fig:maps}
\end{figure}
%%%%%%%%%

\section{Results and Discussion}
\label{sec:res}
Figure \ref{fig:cm_lp} and Table \ref{tab:macro-f1} compare LP performance across ViT architectures initialized either from our FetalUS-188K pretraining or from DINOv3 weights trained on natural images. Models initialized with FetalUS-188K consistently achieve superior classification performance, confirming that exposure to US-specific visual features during pretraining is crucial for effective transfer of frozen representations.
The largest improvement is observed in the classification of the TV plane, an anatomically intermediate view between the TT and TC planes. In ViT-B and ViT-L models pretrained on natural images (Figures \ref{fig:cm_lp}e–f), the three-class problem collapses into a binary decision, with most TV samples misclassified as TT or TC. This reflects the spatial continuity of fetal brain anatomy, where TV shares midline and ventricular features with TT and posterior structures with TC. Without domain-specific pretraining, DINOv3 fails to capture these fine-grained distinctions, relying instead on coarse textural differences that suffice for natural image tasks but not for subtle anatomical transitions.
Interestingly, the ViT-S architecture pretrained on LVD-1689M (Figure \ref{fig:cm_lp}d) outperforms its larger counterparts under LP. This suggests that smaller models, owing to stronger regularization and higher feature compactness, may yield more robust frozen representations when the pretraining domain is suboptimal. 

Conversely, all FetalUS-initialized models (Figures \ref{fig:cm_lp}a–c) show balanced performance across all three classes, highlighting the effectiveness of US-adaptive pretraining in encoding domain-relevant structural cues.
Figure \ref{fig:maps} further illustrates how FetalUS-pretrained DINOv3 focuses on discriminative anatomical landmarks such as the cerebellum, choroid plexus, and ventricular structures. The PCA maps reveal that the model learns feature embeddings that correspond to meaningful anatomical regions, emphasizing areas critical for differentiating between neighboring planes.

While LP performance offers insight into the intrinsic quality of learned representations, full FT allows all parameters to adapt to the target task. As shown in Table \ref{tab:macro-f1-ft}, the weighted F1-scores for the FetalUS-initialized model are significantly higher ($\sim20$\%) compared to the LVD-initialized one. Even with complete parameter optimization, the LVD-initialized model (Figure \ref{fig:cm_all}a) achieves only 45.9\% recall on TV, with confusion spread across classes—indicating that poor initialization hampers convergence despite end-to-end training. In contrast, the FetalUS-initialized model (Figure \ref{fig:cm_all}b) attains 65.0\% TV recall and 84.4\% TT recall, demonstrating that proper pretraining not only yields more discriminative features but also provides a more favorable optimization landscape for FT.

Notably, the performance gap between FetalUS-LP and FetalUS-FT is relatively small compared to the improvement seen in LVD-initialized models after FT. This suggests that FetalUS-pretrained representations already encode most of the information required for accurate plane discrimination, with FT providing only incremental gains. In contrast, natural-image pretraining leads to suboptimal initialization that FT alone cannot overcome.
These findings contrast with recent work \cite{liu2025does} reporting strong DINOv3 transfer to medical imaging tasks, which mainly involved datasets with high inter-class variability (e.g., distinct organs or pathologies). Our study highlights a critical limitation of such generalization: when classes exhibit low inter-class variability and high anatomical overlap, as in fetal brain planes, generic foundation models fail to discriminate without domain adaptation.

Despite these promising results, the FetalUS-188K dataset—while the largest of its kind—may still be too limited to fully explore the scalability of pretraining. Future work will involve stress-testing model robustness to varying dataset sizes and diversities to assess the saturation point of representation quality. Expanding the dataset with more subjects, gestational ages, and US systems could further enhance generalization and reduce potential biases. Additionally, multi-view pretraining or cross-domain adaptation (e.g., from fetal to neonatal brain US) represent promising directions to refine feature transferability across developmental stages and imaging protocols.

%%%%%%%%%%%%%%%%%%%%%%%%%

\section{Conclusion}
\label{sec:concl}
In this work, we conducted the first systematic evaluation of DINOv3 on fetal brain plane classification with low inter-class variability. By aggregating all publicly available fetal US datasets, we established a unified multi-center benchmark for the challenging task of distinguishing anatomically similar planes: TV, TC, and TT views.
Our findings reveal that generic DINOv3 pretrained on natural images fail catastrophically on this fine-grained classification task. 
Domain-adaptive pretraining on fetal US data proves essential, enabling robust classification even with frozen features. These results have important implications for deploying foundation models in settings where accurate plane identification directly impacts diagnostic decisions.

\section*{Acknowledgement}
We acknowledge the CINECA award under the ISCRA initiative, for the availability of high-performance computing resources and support.

\section*{Declaration}
\noindent \textbf{Conflict of interest}
The authors declare that no benefits in any form have been or will be received and there are no conflicts of interest to be disclosed.
\\
\noindent \textbf{Ethics approval}
This study did not need any ethical approval.
\\
\noindent \textbf{Informed consent}
All data used in this study was downloaded from open-source datasets 
\\
\noindent \textbf{Open Access} All data used in this study is openly available online.

\bibliography{sn-bibliography}% common bib file
%% if required, the content of .bbl file can be included here once bbl is generated
%%\input sn-article.bbl

%% Default %%
%%\input sn-sample-bib.tex%

\end{document}